\documentclass[10pt]{article}

\usepackage[margin=1in]{geometry}
\usepackage{times}
\usepackage{graphicx}
\usepackage{booktabs}
\usepackage{amsmath}
\usepackage{amssymb}
\usepackage{enumitem}
\usepackage{xcolor}
\usepackage{algorithm}
\usepackage{algpseudocode}
\usepackage{pgfplots}
\usepackage[hidelinks]{hyperref}
\pgfplotsset{compat=1.18}

\title{Hypergraph Normal World Models for Logical Visual Anomaly Detection}
\author{Weizhi Nie \quad Zibo Xu \quad Weijie Wang \quad Yuting Su\\
Tianjin University}
\date{}

\newcommand{\method}{Hypergraph Normal World Model}
\newcommand{\IQ}{\mathcal{Q}}
\newcommand{\R}{\mathbb{R}}
\definecolor{methodnavy}{HTML}{0B1F4D}
\definecolor{methodblue}{HTML}{2D62AA}
\definecolor{methodteal}{HTML}{00707C}
\definecolor{methodgreen}{HTML}{1F875C}
\definecolor{methodred}{HTML}{C62323}
\definecolor{methodgray}{HTML}{F5F7FA}

\begin{document}
\maketitle

\begin{abstract}
Visual anomaly detection is often deployed with only normal training images.
Most one-class detectors map test patches or features to a normal reference distribution.
This works well for local structural defects.
Logical anomalies are different.
Each visible part may look normal, while the whole image violates a normal count, co-occurrence, or spatial relation.
This paper studies whether a model can learn such a category-specific normal world from nominal images alone.
We propose the \method, a normal-only detector that distills frozen DINOv2 patch tokens into patch, relation, and hypergraph statistics.
It builds spatial hyperedges over token groups.
It then scores each test image with an information quotient that separates local, relational, hyperedge, and hyperedge-relation evidence.
On the available MVTec LOCO breakfast-box validation data, the full hypergraph model improves logical anomaly AUROC from 0.8434 for DINOv2 patch-kNN to 0.9279.
It also improves over the non-hypergraph variant, from 0.9013 to 0.9279.
Few-shot experiments show that the model remains effective with very limited normal images.
We also test whether the score reflects normal-world knowledge rather than a shallow mapping.
t-SNE separates logical anomalies in the learned energy space.
Relation counterfactuals increase the information quotient by 83.13 on average.
Random hypergraphs reduce logical AUROC, and hyperedge attribution is much larger on logical anomalies.
Qualitative examples show that high scores are driven by relation-bearing terms.
These results suggest that logical visual anomaly detection should model normal relations, not only normal local patches.
\end{abstract}

\section{Introduction}
Visual anomaly detection aims to decide whether a test image is consistent with the normal state of an object, product, or visual environment.
In practical inspection systems, the data distribution is highly unbalanced.
Normal samples are easy to collect because they are produced by the regular process.
Abnormal samples are rare, diverse, and often unavailable before deployment.
Even when a few abnormal examples exist, they cannot cover the space of possible failures.
This reality has made one-class, few-shot, and zero-shot anomaly detection an important problem in computer vision.
Most existing methods therefore avoid training a standard supervised classifier.
They instead learn a mapping from an input image, patch, or feature to a normal reference.
This reference may be a nearest-neighbor memory bank, a Gaussian patch distribution, a reconstruction target, a flow likelihood, a teacher-student discrepancy, or a vision-language prompt response~\cite{cohen2020spade,defard2020padim,roth2022patchcore,zavrtanik2021draem,yu2021fastflow,deng2022reverse,jeong2023winclip,batzner2024efficientad,zhou2024anomalyclip}.
This line of work has produced strong detectors, especially when the anomaly creates visible local evidence such as a scratch, stain, missing texture, or broken surface.

However, the mapping view still has an important limitation.
It asks whether a test image can be matched to, reconstructed from, or classified against a set of learned normal references.
When the number of normal images is small, this reference set is incomplete.
When the abnormal space is open ended, no mapping can cover all invalid cases.
More importantly, some failures are not failures of local appearance.
They are failures of the visual world's internal rules.
An image may be composed of normal-looking parts, but the parts may have a wrong count, a wrong co-occurrence pattern, an impossible order, or an inconsistent spatial relation.
MVTec LOCO AD was designed to expose this issue by separating logical anomalies from structural anomalies~\cite{bergmann2022loco}.
In this setting, the key question is not only whether each patch has a familiar appearance.
The key question is whether the image can be explained as a valid normal world.

This paper addresses the problem of \emph{logical} visual anomaly detection from normal images only.
We study whether a model can learn the normal world of one visual category without abnormal examples.
We also test whether the learned world can reject logically invalid images.
The problem is difficult for two reasons.
\begin{itemize}[leftmargin=*]
    \item \textbf{Challenge 1. Few normal samples provide incomplete coverage.}
    In a normal-only setting, the model cannot rely on abnormal labels.
    In a few-shot setting, it also lacks dense coverage of normal appearances.
    A direct mapping from test patches to stored normal patches is therefore fragile.
    It can memorize the observed normal samples, but it may fail to describe the latent rule that makes those samples normal.
    A detector needs a prior that is broader than the target training set, while still being calibrated to the target category.
    \item \textbf{Challenge 2. Single-category normality is relational, not only local.}
    Logical anomalies are created by invalid configurations of normal parts.
    A detector must learn what is essential in the normal category.
    It must know which regions should appear, which parts should co-occur, which groups should be compatible, and which relations should remain stable.
    This knowledge cannot be evaluated only by a local distance to the nearest normal patch.
    The score should measure how much information is required to explain the whole image under the learned normal world.
\end{itemize}

To address the first challenge, we use a visual foundation model as a source of world knowledge and distill a category-specific normal world from it.
A large self-supervised model has already learned broad visual regularities from massive unlabeled data.
Its patch tokens contain semantic cues about objects, parts, and layouts that cannot be learned from a few target images alone.
We therefore do not train a visual encoder from scratch.
Instead, we freeze the encoder and use the available normal images to estimate what the target category looks like inside this pretrained feature space.
This gives the model a low-shot prior and keeps the normal-only protocol.

\begin{figure}[t]
    \centering
    \includegraphics[width=\linewidth]{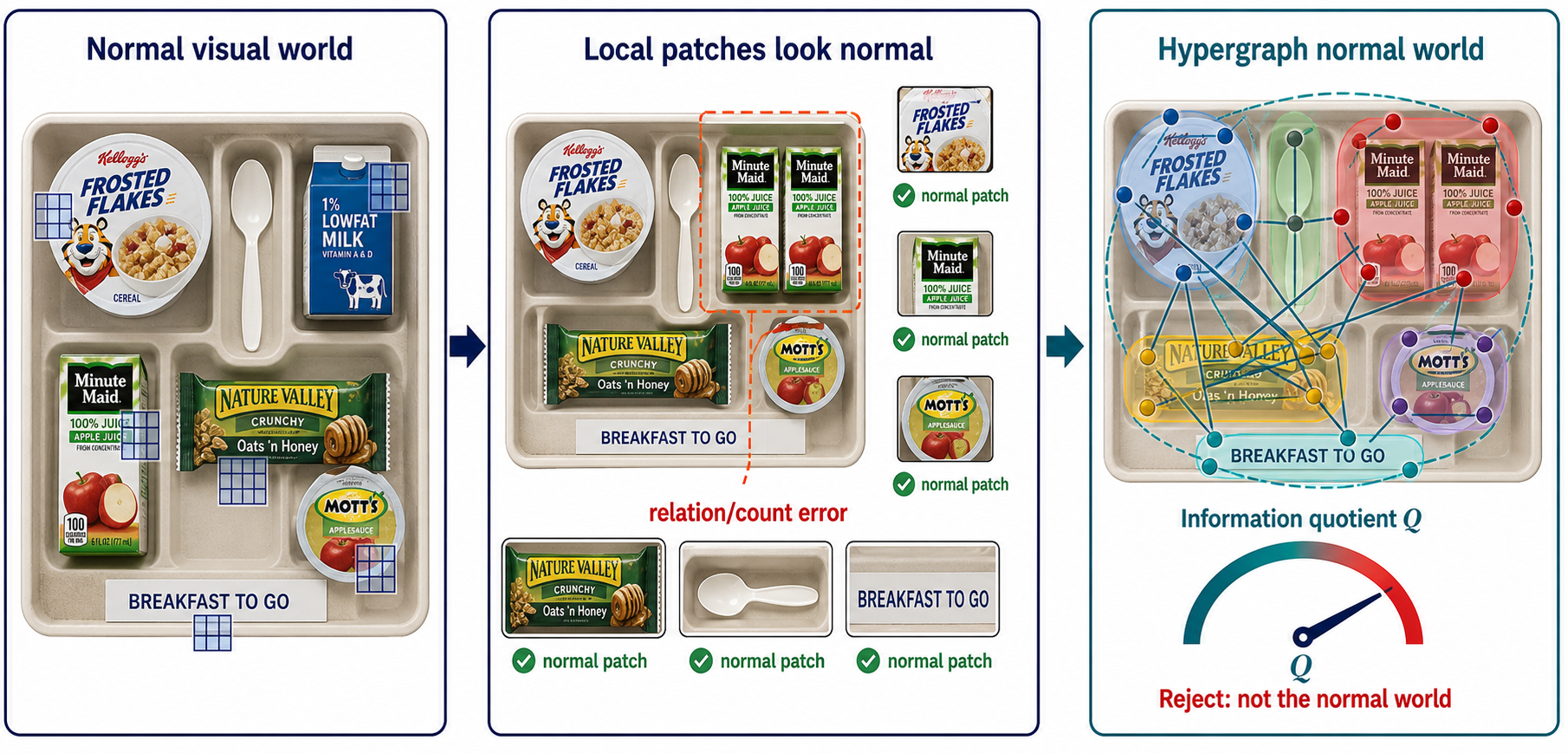}
    \caption{Motivation.
    Patch-memory baselines are strong local normality models.
    Logical anomalies can preserve local appearance while breaking relations among parts.
    We model the normal visual world with hyperedges over patch tokens and use an information quotient to measure relation violations.}
    \label{fig:motivation}
\end{figure}

To address the second challenge, we treat a normal image as a structured world rather than a bag of independent patches.
The normal category is represented by local tokens, larger region groups, and relations among these groups.
If a test image is normal, these elements should be easy to explain under the statistics learned from normal images.
If the image is logically abnormal, the local tokens may still be familiar, but the relational structure should become inefficient to encode.
We capture this idea with an information quotient.
The quotient measures the relative explanation cost of a test image under the normal world.
It is high when patch groups or cross-region relations deviate from the normal relational code.
This can happen even when many individual patches remain visually plausible.
Figure~\ref{fig:motivation} illustrates this motivation.

Based on these ideas, we propose the \method.
The method is a normal-only anomaly detector built on frozen DINOv2 patch tokens~\cite{oquab2023dinov2}.
It turns the conceptual goal of learning a normal world into a concrete scoring pipeline.
\begin{itemize}[leftmargin=*]
    \item \textbf{First, we distill general visual knowledge into target normal tokens.}
    The model extracts a grid of semantic patch tokens from each image with a frozen DINOv2 encoder.
    The encoder is not updated on the target data.
    The target normal images are used only to calibrate normal-world statistics in the pretrained feature space.
    This step is important in zero-shot and few-shot regimes.
    The model can use broad visual knowledge from large-scale pretraining without overfitting to a tiny normal set.
    \item \textbf{Second, we organize the token grid as a hypergraph normal world.}
    The model constructs fixed spatial hyperedges over the token grid.
    Each hyperedge groups multiple locations, such as a quadrant, center region, border region, row band, column band, or local grid cell.
    This gives the detector a middle level between individual patches and the whole image.
    The model can ask whether a patch group behaves like a normal part.
    It can also ask whether different groups form a valid normal configuration.
    \item \textbf{Third, we score the image by an information quotient.}
    The model estimates normal statistics for four forms of evidence: patch-location evidence, pairwise patch-relation evidence, hyperedge evidence, and hyperedge-relation evidence.
    At inference time, these energies are combined into an information quotient $\IQ$.
    A low quotient means that the image is efficiently explained by the normal world.
    A high quotient means that the image requires abnormal relational information and should be rejected.
\end{itemize}

The experiments support this formulation.
On the available MVTec LOCO breakfast-box validation data, the hypergraph normal-world model improves logical anomaly AUROC from 0.8434 for a DINOv2 patch-kNN baseline to 0.9279.
Patch-kNN remains stronger for structural anomalies, which is consistent with its local-memory design.
The proposed model is stronger when the abnormality is relational.
Additional analysis gives the same message.
Random hypergraphs weaken logical detection.
Relation counterfactuals sharply increase the information quotient.
Hyperedge attribution is much larger on logical anomalies than on normal images.

Our contributions are:
\begin{itemize}[leftmargin=*]
    \item We formulate logical visual anomaly detection as normal-world modeling over frozen foundation-model patch tokens.
    The formulation addresses few-shot normal-only learning by distilling category-specific normality from a pretrained visual world model.
    \item We introduce a hypergraph information quotient that combines patch evidence, pairwise relation evidence, hyperedge evidence, and hyperedge-relation evidence.
    The score evaluates whether a test image is efficient to explain under learned normal-world structure.
    \item We evaluate the method with baseline comparisons, module ablations, few-shot studies, sensitivity analysis, and five interpretability experiments.
    The results show a clear advantage on logical anomalies.
    They also clarify the complementary behavior between relation-aware scoring and patch-kNN.
\end{itemize}

\section{Related Work}
\paragraph{Benchmarks and the normal-only inspection protocol.}
Industrial anomaly detection is usually evaluated under a normal-only training protocol.
MVTec AD established a widely used real-world benchmark with object and texture categories, normal training images, and anomalous test images~\cite{bergmann2019mvtec}.
This benchmark made it possible to compare one-class detectors without requiring defect labels during training.
However, many categories in MVTec AD are dominated by structural defects, where the abnormal evidence is local.
MVTec LOCO AD extends this setting.
It introduces logical constraint anomalies and reports logical and structural subsets separately~\cite{bergmann2022loco}.
This separation is central to our paper.
It turns the question from ``can the model find a defective patch'' into ``can the model reject a visually plausible but logically invalid world''.
Recent work such as EfficientAD also recognizes the importance of logical anomalies.
It adds a global autoencoder branch to handle invalid combinations of normal local features~\cite{batzner2024efficientad}.
Our work follows the same motivation but uses an explicit normal-world relation model rather than only a global reconstruction branch.

\paragraph{Patch distributions, memory banks, and target-adapted features.}
A large family of strong anomaly detectors represents each image by local features and compares test features to nominal feature statistics.
SPADE performs anomaly localization by matching deep pyramid features from a test image to normal images~\cite{cohen2020spade}.
PaDiM models pretrained patch features with location-wise Gaussian statistics~\cite{defard2020padim}.
PatchCore stores a compact coreset of nominal patch features and scores each test image by nearest-neighbor distances in patch space~\cite{roth2022patchcore}.
CFA adapts features to the target category with a coupled-hypersphere objective and a scalable memory bank~\cite{lee2022cfa}.
ReConPatch further learns contrastive patch representations from nominal data to make target features more discriminative~\cite{hyun2024reconpatch}.
These methods are powerful because local pretrained features carry strong visual semantics and because many industrial defects are local distribution shifts.
They are also the right baselines for our question.
If a logical anomaly is visible as a collection of unusual local features, patch-memory models should perform well.
Our method targets the remaining case.
The local features are familiar, but their relations are not.

\paragraph{Reconstruction, synthetic anomalies, flows, and distillation.}
Another line of work defines normality through reconstruction or prediction.
DFR reconstructs deep regional features for unsupervised anomaly segmentation~\cite{yang2020dfr}.
DRAEM trains a discriminative reconstruction embedding using synthetic anomaly generation~\cite{zavrtanik2021draem}.
CutPaste creates self-supervised anomalies by cutting and pasting image patches, then learns representations that separate original and modified images~\cite{li2021cutpaste}.
Flow-based methods such as CFLOW-AD and FastFlow estimate feature likelihoods with conditional or two-dimensional normalizing flows~\cite{gudovskiy2022cflow,yu2021fastflow}.
Student-teacher methods train a student network to match a pretrained teacher on normal images and use prediction discrepancy as the anomaly signal~\cite{wang2021stfpm,deng2022reverse,batzner2024efficientad}.
These approaches provide different normality criteria: reconstruction error, likelihood, or distillation mismatch.
They often model dense local deviations well, and some include global branches.
Our method is complementary.
It does not train a new image generator or student network.
It estimates relation statistics over frozen tokens.
It then asks whether the test image is efficient to explain under a normal relational code.

\paragraph{Foundation models and zero-/few-shot anomaly detection.}
Large pretrained visual models have changed the low-shot anomaly detection setting.
ResNet features remain a common transfer-learning backbone~\cite{he2016resnet}, while Vision Transformers expose images as patch-token sequences~\cite{dosovitskiy2021vit}.
Self-supervised ViTs such as DINO show that patch tokens can contain semantic and object-level information without labels~\cite{caron2021dino}.
Masked autoencoders further demonstrate that scalable self-supervised pretraining can produce transferable visual representations from patch reconstruction objectives~\cite{he2022mae}.
DINOv2 improves this direction with robust visual features learned without supervision~\cite{oquab2023dinov2}.
Vision-language models add another route.
CLIP provides transferable image-text representations~\cite{radford2021clip}, and WinCLIP adapts CLIP to zero- and few-shot anomaly classification and segmentation with window-level features and prompt ensembles~\cite{jeong2023winclip}.
AnomalyCLIP learns object-agnostic prompts for zero-shot anomaly detection across domains~\cite{zhou2024anomalyclip}.
These works show that large pretrained models can provide useful priors when target data are scarce.
Our method uses this idea differently.
We do not rely on text prompts or train the visual encoder.
Instead, we distill a category-specific normal world from frozen DINOv2 patch tokens using only normal images.

\paragraph{Relation-aware normality.}
The above methods show strong progress, but most of them still score an image by local feature rarity, reconstruction quality, likelihood, or teacher-student mismatch.
Logical anomaly detection requires a more relational notion of normality.
The detector must compare groups of regions and decide whether their joint configuration is compatible with nominal examples.
Our hypergraph construction is a simple way to impose this middle level.
Patch tokens model local evidence, hyperedges model region groups, and hyperedge relations model interactions between groups.
This design connects foundation-model features with a normal-only structural score.
It also gives interpretable evidence.
The information quotient can be decomposed into patch, relation, hyperedge, and hyperedge-relation terms.
The most responsible hyperedges can then be inspected in analysis.

\paragraph{Broader visual domain knowledge and structured priors.}
Several related studies outside standard industrial anomaly detection also motivate our design choices.
Multigranular visual-semantic embedding has been used to represent identity under appearance changes~\cite{gao2022multigranular}.
Transformer-based prior knowledge has also been used to improve low-level visual enhancement~\cite{wen2023mrft}.
Unknown fault detection has been studied through similarity mining of stationary and non-stationary features~\cite{li2023unknownfault}.
Graph convolution has also been used for ship behavior anomaly detection~\cite{ma2026mfgcn}.
Self-supervised cross-domain medical segmentation and depth-aware scene understanding show the value of structured visual priors under limited labels~\cite{li2025causalfundus,chen2025dsdp}.
Zero-shot cross-modal retrieval and vision-language prior modeling further support the transfer of pretrained semantic structure to low-shot visual tasks~\cite{li2026cdgan,liu2026causalcompnet}.
These works are not direct baselines for MVTec LOCO AD, but they support the broader direction of using structured visual knowledge rather than relying only on direct sample-to-label mapping.

\section{Method}
\subsection{Problem Definition}
Let $\mathcal{D}_{n}=\{x_i\}_{i=1}^{N}$ be the normal training images of one visual category.
No anomalous image and no anomaly mask are used during fitting.
Given a test image $x$, the model outputs an anomaly score $\IQ(x)$.
The image is accepted as normal when $\IQ(x)\leq\tau$ and rejected as anomalous when $\IQ(x)>\tau$.
The threshold $\tau$ is calibrated from normal scores.
In the experiments, we use the 95th percentile of normal calibration scores.

The goal is to learn a category-specific normal world from $\mathcal{D}_{n}$.
This normal world should not only describe local appearance.
It should also describe which regions co-occur and which region groups have stable relations.
We denote the fitted normal-world parameters by
\begin{equation}
    \Theta_{\mathrm{NW}} =
    \{\Theta_{\mathrm{cls}},\Theta_{\mathrm{patch}},
      \Theta_{\mathrm{rel}},\Theta_{\mathrm{hist}},
      \Theta_{\mathrm{hyper}},\Theta_{\mathrm{hrel}}\}.
\end{equation}
Each component stores normal means, variances, and calibration statistics for one type of evidence.

\subsection{Overview}
\begin{figure}[t]
    \centering
    \includegraphics[width=\linewidth]{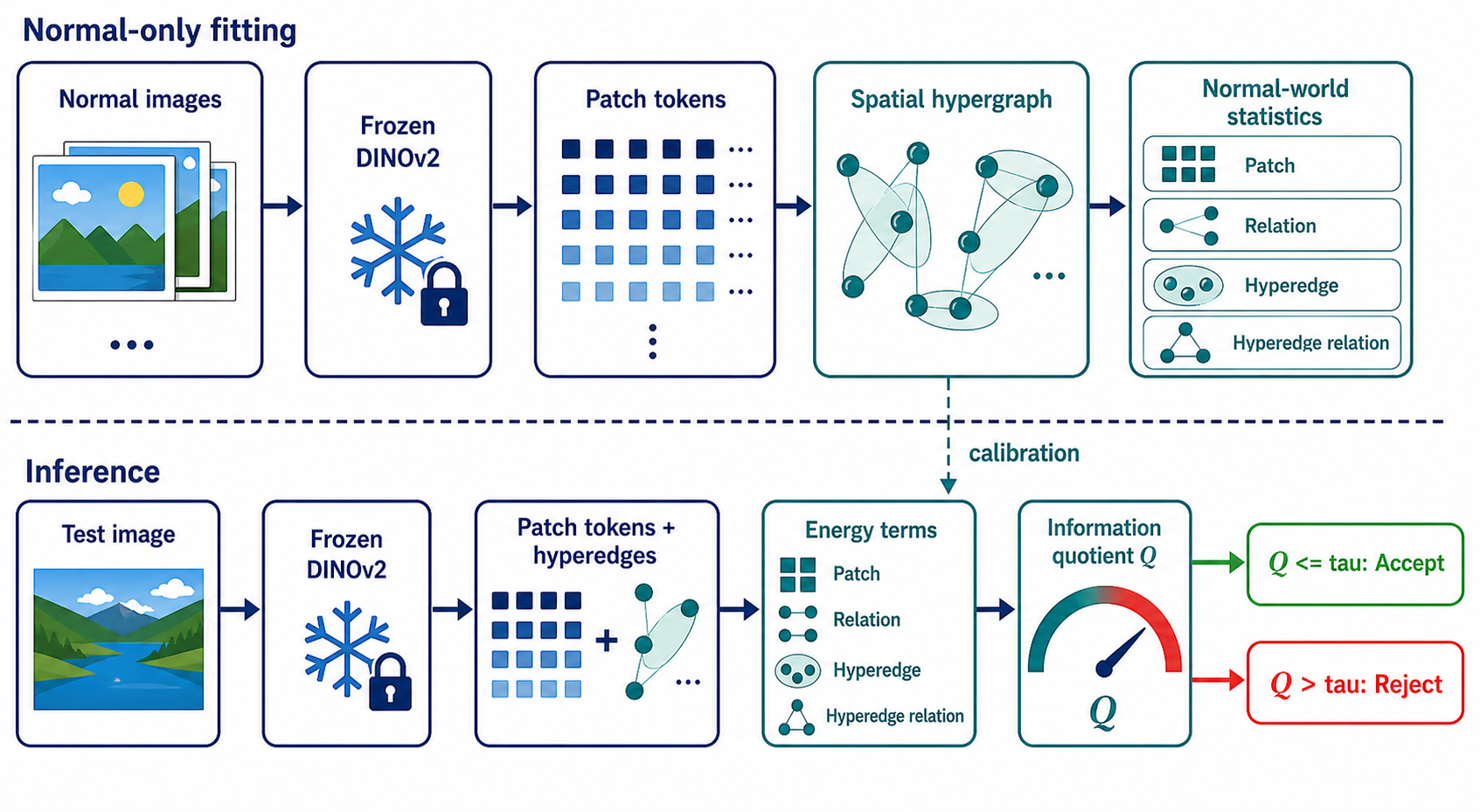}
    \caption{Framework of the proposed hypergraph normal-world model.
    Normal images are used to estimate token, relation, hyperedge, and hyperedge-relation statistics in a frozen DINOv2 feature space.
    A test image is scored by standardized normal-world energies.
    The information quotient $\IQ$ is the final anomaly score.}
    \label{fig:framework}
\end{figure}

Figure~\ref{fig:framework} gives the full pipeline.
The model has two stages.
During normal-only fitting, a frozen vision foundation model maps every normal image to semantic tokens.
A fixed hypergraph groups these tokens into spatial regions.
The method then estimates normal statistics for tokens, relations, hyperedges, and hyperedge relations.
During inference, the same feature and hypergraph construction is applied to a test image.
The test image receives a set of normal-world energies.
The standardized weighted sum of these energies is the information quotient $\IQ$.

\subsection{Frozen Token Representation}
We use a frozen DINOv2 encoder $f$ as the visual world prior.
For an image $x$, the encoder outputs a class token $c(x)\in\R^d$ and a grid of patch tokens
\begin{equation}
    Z(x)=\{z_1(x),\ldots,z_T(x)\}, \quad z_t(x)\in\R^d .
\end{equation}
For the ViT-B/14 teacher and $224\times224$ inputs, $T=16\times16$.
The teacher is fixed.
Only statistics over its output tokens are fitted on the target normal images.
This design separates general visual knowledge from category-specific normal calibration.

\subsection{Hypergraph Normal World}
We define a fixed hypergraph $\mathcal{G}=(\mathcal{V},\mathcal{E}_h)$ on the patch grid.
The node set $\mathcal{V}=\{1,\ldots,T\}$ contains patch locations.
Each hyperedge $e_m\in\mathcal{E}_h$ is a subset of patch locations, and $M=|\mathcal{E}_h|$.
The implementation uses several spatial groups: quadrants, center and border regions, row bands, column bands, and a $4\times4$ local grid.
For each hyperedge, we compute a pooled hyperedge token
\begin{equation}
    h_m(x)=\frac{1}{|e_m|}\sum_{t\in e_m} z_t(x).
\end{equation}
The hyperedge token is a region-level state.
It is more stable than an individual patch and more local than the whole-image representation.
It allows the model to represent normal parts and relations among normal parts.

We also compute relation states.
For selected patch-location pairs $(u,v)\in\mathcal{P}$, the patch relation is
\begin{equation}
    r^{p}_{uv}(x)=
    \operatorname{cos}\!\left(z_u(x),z_v(x)\right).
\end{equation}
For selected hyperedge pairs $(m,n)\in\mathcal{P}_h$, the hyperedge relation is
\begin{equation}
    r^{h}_{mn}(x)=
    \operatorname{cos}\!\left(h_m(x),h_n(x)\right).
\end{equation}
These relation states are the core difference from a pure patch-memory model.
They ask whether regions agree with each other in the same way that normal regions agree.

\subsection{Normal-World Statistics}
From normal images, we estimate diagonal Gaussian statistics for each state.
For the class token, we store $\mu_c$ and $\sigma_c$.
For each patch location $t$, we store $\mu_t$ and $\sigma_t$.
For each hyperedge $m$, we store $\mu^h_m$ and $\sigma^h_m$.
For each selected patch relation $(u,v)$ and hyperedge relation $(m,n)$, we store scalar means and standard deviations:
\begin{equation}
    \mu^{p}_{uv},\sigma^{p}_{uv}
    \quad\text{and}\quad
    \mu^{hrel}_{mn},\sigma^{hrel}_{mn}.
\end{equation}
We also compute a relation histogram $g(x)$ over patch-pair similarities and store its normal mean $\mu_g$ and standard deviation $\sigma_g$.
The histogram gives a global summary of the relation distribution.
The relation matrix keeps location-specific information.

\subsection{Energy Terms}
Each component measures how much information is needed to explain one aspect of $x$ under the normal world.
The class-token energy is
\begin{equation}
    E_{\mathrm{cls}}(x)=
    \frac{1}{d}
    \left\|
        \frac{c(x)-\mu_c}{\sigma_c+\epsilon}
    \right\|_2^2 .
\end{equation}
This term captures coarse image-level compatibility.
It is useful when the whole image is far from the normal category, but it is too coarse for local or logical diagnosis.

The patch-location energy is
\begin{equation}
    E_{\mathrm{patch}}(x)=
    \frac{1}{Td}\sum_{t=1}^{T}
    \left\|
        \frac{z_t(x)-\mu_t}{\sigma_t+\epsilon}
    \right\|_2^2 .
\end{equation}
This term measures whether each spatial token looks normal at its location.
It is closest to a location-aware patch distribution model.

The patch-relation energy is
\begin{equation}
    E_{\mathrm{rel}}(x)=
    \frac{1}{|\mathcal{P}|}\sum_{(u,v)\in\mathcal{P}}
    \left(
        \frac{r^{p}_{uv}(x)-\mu^{p}_{uv}}
             {\sigma^{p}_{uv}+\epsilon}
    \right)^2 .
\end{equation}
This term measures whether patch pairs keep normal similarities.
It can increase when two normal-looking regions appear in an abnormal relation.

The relation-histogram energy is
\begin{equation}
    E_{\mathrm{hist}}(x)=
    \frac{1}{B}
    \left\|
        \frac{g(x)-\mu_g}{\sigma_g+\epsilon}
    \right\|_2^2 ,
\end{equation}
where $B$ is the number of histogram bins.
This term captures the global distribution of pairwise relations.
It is less location-specific than $E_{\mathrm{rel}}$ but more robust to small spatial shifts.

The hyperedge-token energy is
\begin{equation}
    E_{\mathrm{hyper}}(x)=
    \frac{1}{Md}\sum_{m=1}^{M}
    \left\|
        \frac{h_m(x)-\mu^h_m}{\sigma^h_m+\epsilon}
    \right\|_2^2 .
\end{equation}
This term evaluates whether a group of patches behaves like a normal region.
It is designed to bridge local patches and the full image.

Finally, the hyperedge-relation energy is
\begin{equation}
    E_{\mathrm{hrel}}(x)=
    \frac{1}{|\mathcal{P}_h|}\sum_{(m,n)\in\mathcal{P}_h}
    \left(
        \frac{r^{h}_{mn}(x)-\mu^{hrel}_{mn}}
             {\sigma^{hrel}_{mn}+\epsilon}
    \right)^2 .
\end{equation}
This term evaluates whether region groups form normal cross-region relations.
It is the most direct term for logical anomalies.

\subsection{Information Quotient and Overall Loss}
The method does not optimize a neural network with backpropagation.
The normal-only fitting stage estimates $\Theta_{\mathrm{NW}}$.
At test time, we use an overall normal-world loss to score the image.
Let
\begin{equation}
    \mathcal{K}=
    \{\mathrm{cls},\mathrm{patch},\mathrm{rel},
      \mathrm{hist},\mathrm{hyper},\mathrm{hrel}\}
\end{equation}
be the set of energy components.
For each component $k\in\mathcal{K}$, we compute the normal calibration mean $m_k$ and standard deviation $s_k$ from normal training or calibration images.
The standardized component loss is
\begin{equation}
    q_k(x)=
    \frac{E_k(x)-m_k}{s_k+\epsilon}.
\end{equation}
The overall normal-world loss is
\begin{equation}
    \mathcal{L}_{\mathrm{NW}}(x)
    = \sum_{k\in\mathcal{K}} \lambda_k q_k(x),
    \qquad
    \lambda_k\geq0,\quad
    \sum_{k\in\mathcal{K}}\lambda_k=1 .
\end{equation}
We define the information quotient as this calibrated loss:
\begin{equation}
    \IQ(x)=\mathcal{L}_{\mathrm{NW}}(x).
\end{equation}
The quotient has a simple interpretation.
If the image follows the normal world, all standardized losses remain close to the normal calibration range.
If the image breaks normal appearance or normal relations, one or more terms become large and $\IQ$ increases.
In the full hypergraph model, all components are used after calibration.
In the ablation study, we switch components on and off.
This tests which part of the normal world matters.

\subsection{Algorithm}
Algorithm~\ref{alg:hnwm} summarizes fitting and inference.
The training stage is statistical fitting rather than gradient optimization.
The inference stage is deterministic once $\Theta_{\mathrm{NW}}$ and $\tau$ are fixed.

\begin{algorithm}[t]
\caption{Hypergraph Normal World Model}
\label{alg:hnwm}
\begin{algorithmic}[1]
\Require Normal images $\mathcal{D}_{n}$, frozen encoder $f$, fixed hypergraph $\mathcal{G}$, weights $\{\lambda_k\}$.
\Ensure Normal-world parameters $\Theta_{\mathrm{NW}}$, threshold $\tau$, and test score $\IQ(x)$.
\State Build hyperedges $\mathcal{E}_h$ on the patch grid.
\For{each normal image $x_i\in\mathcal{D}_{n}$}
    \State Extract $c(x_i),Z(x_i)=f(x_i)$.
    \State Pool hyperedge tokens $\{h_m(x_i)\}_{m=1}^{M}$.
    \State Compute patch relations $\{r^p_{uv}(x_i)\}_{(u,v)\in\mathcal{P}}$.
    \State Compute hyperedge relations $\{r^h_{mn}(x_i)\}_{(m,n)\in\mathcal{P}_h}$.
\EndFor
\State Estimate normal statistics $\Theta_{\mathrm{NW}}$ for tokens, relations, histograms, hyperedges, and hyperedge relations.
\State Compute each normal energy $E_k(x_i)$ and calibration statistics $(m_k,s_k)$.
\State Set threshold $\tau$ as a high normal-score quantile.
\Statex
\Function{Score}{$x$}
    \State Extract $c(x),Z(x)$ with the same frozen encoder.
    \State Build $h_m(x)$, $r^p_{uv}(x)$, $g(x)$, and $r^h_{mn}(x)$.
    \For{each component $k\in\mathcal{K}$}
        \State Compute raw energy $E_k(x)$ and standardized loss $q_k(x)$.
    \EndFor
    \State $\IQ(x)\gets\sum_{k\in\mathcal{K}}\lambda_k q_k(x)$.
    \State \Return $\IQ(x)$ and reject if $\IQ(x)>\tau$.
\EndFunction
\end{algorithmic}
\end{algorithm}

\section{Experiments}
\subsection{Dataset, Baselines, Metrics, and Implementation}
We evaluate on MVTec LOCO AD, a benchmark designed for logical constraint anomaly detection~\cite{bergmann2022loco}.
MVTec LOCO extends the normal-only industrial inspection protocol of MVTec AD~\cite{bergmann2019mvtec} by separating \emph{logical} anomalies from \emph{structural} anomalies.
This separation is important for our goal.
Structural anomalies usually create local visual defects, while logical anomalies may preserve local appearance and only violate object-level or relation-level rules.

The current extracted project data contain the breakfast-box category.
The local manifest has 351 normal training images, 62 normal validation images, 102 normal test images, 83 logical anomalies, and 90 structural anomalies.
All models are fitted with normal training images only.
No anomalous image is used to tune the normal-world statistics.
The validation split is used only for normal calibration and threshold selection.
Only one LOCO category is available in the local project data.
The experiments should therefore be read as a focused validation of the idea, not as a complete five-category benchmark claim.

\paragraph{Baselines.}
We compare against three representative baselines.
\textbf{ResNet50 global feature} uses an ImageNet-pretrained ResNet50 representation~\cite{he2016resnet}.
For each image, we extract a global feature and score the distance to the normal feature distribution.
This baseline tests whether ordinary global transfer features are sufficient for the task.
\textbf{DINOv2 CLS feature} uses the frozen DINOv2 class token~\cite{oquab2023dinov2}.
It is a stronger foundation-model global baseline and tests whether large self-supervised pretraining alone can solve logical anomaly detection.
\textbf{DINOv2 patch-kNN} uses DINOv2 patch tokens and nearest-neighbor distances to normal patch memories.
It follows the central idea of PatchCore-style memory methods~\cite{roth2022patchcore}.
A test image is anomalous when its patches are far from nominal patches.
This is the strongest local-memory baseline in our reproduced setting.

\paragraph{Our methods.}
\textbf{Ours w/o hypergraph} uses the information quotient without hyperedge-level grouping.
It keeps the frozen DINOv2 features and normal-world scoring idea, but removes the explicit hypergraph structure.
\textbf{Ours + hypergraph} is the full proposed model.
It combines patch, relation, hyperedge, and hyperedge-relation evidence in the calibrated normal-world loss.
This comparison isolates whether the hypergraph structure adds useful information beyond the basic normal-world score.

\paragraph{Metrics and implementation.}
We report image-level AUROC and AUPRC over all test images.
We also report AUROC separately on logical anomalies and structural anomalies.
The separated AUROC values are computed by comparing each anomaly group against normal test images.
This makes the failure mode visible.
A method may be strong on structural defects but weak on logical violations, or vice versa.
All methods use the same train/test split and the same normal-only fitting protocol.

\subsection{Comparison with SOTA Baselines}
Table~\ref{tab:main} gives the main comparison with the reproduced SOTA-style baselines.
The global ResNet50 baseline performs the weakest.
Its all-sample AUROC is 0.6554 and its logical AUROC is 0.7049.
This result is expected.
A single global feature can capture coarse category mismatch, but it loses many local and relational details.

Replacing ResNet50 with the DINOv2 class token gives a clear gain.
DINOv2 CLS improves all-sample AUROC from 0.6554 to 0.7635 and logical AUROC from 0.7049 to 0.8616.
This shows that foundation-model pretraining provides a much stronger visual prior.
However, the structural AUROC remains 0.6730.
A global token is still not enough for dense local defect evidence.

DINOv2 patch-kNN is the strongest baseline overall.
It reaches the best all-sample AUROC, the best AUPRC, and the best structural AUROC.
This confirms the value of local patch memory for structural defects.
At the same time, its logical AUROC is 0.8434, which is lower than the DINOv2 CLS baseline and much lower than our full model.
This is the key observation behind our method.
A local patch can be familiar even when the whole configuration is invalid.

Our normal-world score without hypergraph already improves logical AUROC to 0.9013.
This suggests that calibrated normal-world energies capture more than nearest-neighbor patch rarity.
Adding the hypergraph further raises logical AUROC to 0.9279.
The full model does not beat patch-kNN on structural defects, but it is clearly stronger on logical anomalies.
The conclusion is therefore not that hypergraph normal-world modeling replaces patch memory in every case.
The conclusion is that it supplies the missing relation-aware signal for logical anomaly detection.

\begin{table}[t]
\centering
\caption{Main comparison on the available MVTec LOCO breakfast-box data.}
\label{tab:main}
\begin{tabular}{lrrrr}
\toprule
Method & All AUROC & All AUPRC & Logical AUROC & Structural AUROC \\
\midrule
ResNet50 global feature & 0.6554 & 0.7776 & 0.7049 & 0.6098 \\
DINOv2 CLS feature & 0.7635 & 0.8676 & 0.8616 & 0.6730 \\
DINOv2 patch-kNN & \textbf{0.8546} & \textbf{0.9253} & 0.8434 & \textbf{0.8650} \\
Ours w/o hypergraph & 0.8022 & 0.8914 & 0.9013 & 0.7108 \\
Ours + hypergraph & 0.8201 & 0.9043 & \textbf{0.9279} & 0.7207 \\
\bottomrule
\end{tabular}
\end{table}

\subsection{Ablation Study}
Table~\ref{tab:ablation} isolates the contribution of each normal-world component.
The ablation answers two questions.
First, is the improvement only caused by using DINOv2 features?
Second, does the hypergraph structure contribute beyond ordinary patch and relation statistics?

The manifold-only baseline reaches 0.8616 logical AUROC, which is similar to the DINOv2 CLS result in Table~\ref{tab:main}.
This shows that global foundation features are helpful but incomplete.
The patch-grid-only variant reaches 0.8486 logical AUROC and 0.7046 structural AUROC.
It preserves location-aware token statistics, but it still lacks an explicit mechanism for grouping regions.
The graph relation matrix and relation histogram variants are weaker when used alone.
This indicates that pairwise relations are useful evidence.
Raw relations alone are not stable enough to define the whole normal world.

The hyperedge node energy is the strongest single component, reaching 0.9063 logical AUROC.
This is an important result.
It means that grouping patches into normal region states already captures much of the logical structure.
Hyperedge relation alone is weaker than hyperedge node energy, but it still provides complementary cross-region evidence.
The full hypergraph model combines the stable region representation and the relation-level score.
It achieves the best all-sample, logical, and structural AUROC among our variants.
This validates the design choice of using a multi-level normal world rather than a single energy term.

\begin{table}[t]
\centering
\caption{Ablation study of normal-world energy components.}
\label{tab:ablation}
\begin{tabular}{lrrr}
\toprule
Variant & All AUROC & Logical AUROC & Structural AUROC \\
\midrule
Ours w/o hypergraph & 0.8022 & 0.9013 & 0.7108 \\
Manifold only & 0.7635 & 0.8616 & 0.6730 \\
Patch-grid only & 0.7737 & 0.8486 & 0.7046 \\
Graph relation matrix only & 0.7538 & 0.8428 & 0.6717 \\
Relation histogram only & 0.7344 & 0.8434 & 0.6340 \\
Hyperedge node only & 0.8012 & 0.9063 & 0.7042 \\
Hyperedge relation only & 0.7714 & 0.8599 & 0.6899 \\
Full hypergraph model & \textbf{0.8201} & \textbf{0.9279} & \textbf{0.7207} \\
\bottomrule
\end{tabular}
\end{table}

\subsection{Few-Shot Normal-Only Fitting}
Table~\ref{tab:fewshot} and Figure~\ref{fig:fewshot} study how many normal images are needed to fit the normal world.
For 1, 2, 4, 8, and 16 normal images, we average over three random seeds.
The full-data result uses all 351 normal training images.

The model is already meaningful with one normal image.
It reaches 0.8597 logical AUROC, which is close to the DINOv2 patch-kNN baseline in Table~\ref{tab:main}.
This supports the use of a frozen foundation model as a visual prior.
The target data do not need to teach the model what general objects and parts are; they only calibrate the normal-world statistics for the target category.
As more normal images are added, logical AUROC rises from 0.8597 with one image to 0.9194 with 16 images.
It reaches 0.9279 with all 351 images.
The saturation trend is useful in practice.
It suggests that the method can work in a few-shot setting, while still benefiting from more complete normal coverage.

The structural AUROC grows more slowly.
It increases from 0.6167 with one normal image to 0.7207 with the full training set.
This again matches the main comparison.
Our method is designed to model logical structure, not to be the best local defect detector.
For deployment, this suggests a natural hybrid direction.
Patch memory can handle structural defects, while the hypergraph information quotient can handle logical violations.

\begin{table}[t]
\centering
\caption{Few-shot results for the hypergraph model.}
\label{tab:fewshot}
\begin{tabular}{rrrr}
\toprule
Normal training images & All AUROC & Logical AUROC & Structural AUROC \\
\midrule
1 & 0.7333 & 0.8597 & 0.6167 \\
2 & 0.7346 & 0.8473 & 0.6307 \\
4 & 0.7852 & 0.8831 & 0.6949 \\
8 & 0.7902 & 0.9078 & 0.6817 \\
16 & 0.8093 & 0.9194 & 0.7078 \\
351 & \textbf{0.8201} & \textbf{0.9279} & \textbf{0.7207} \\
\bottomrule
\end{tabular}
\end{table}

\begin{figure}[t]
\centering
\caption{Few-shot normal-only fitting.
Bars show logical AUROC as the number of normal training images increases.}
\label{fig:fewshot}
\begin{tikzpicture}
\begin{axis}[
    ybar,
    width=0.82\linewidth,
    height=5.2cm,
    ymin=0.82,
    ymax=0.95,
    ylabel={Logical AUROC},
    xlabel={Number of normal training images},
    symbolic x coords={1,2,4,8,16,351},
    xtick=data,
    bar width=12pt,
    ymajorgrids=true,
    grid style={draw=gray!20},
    nodes near coords,
    every node near coord/.append style={font=\scriptsize, rotate=90, anchor=west},
    tick label style={font=\small},
    label style={font=\small},
]
\addplot[fill=methodteal!65, draw=methodteal] coordinates {
    (1,0.8597) (2,0.8473) (4,0.8831) (8,0.9078) (16,0.9194) (351,0.9279)
};
\end{axis}
\end{tikzpicture}
\end{figure}

\section{Does the Model Learn a Normal World?}
The central claim of this paper is stronger than a metric improvement.
We do not only want a detector that maps a test image to an anomaly score.
We want evidence that the score reflects learned normal-world knowledge.
In particular, the model should know that normal-looking parts must appear in a normal configuration.
If the method only learns a shallow mapping from appearance to score, changing relations should not produce a systematic and interpretable change.
If the method learns a normal world, relation-breaking interventions should affect the information quotient in predictable ways.
The same should hold for hypergraph structure and hyperedge attribution.

We therefore analyze the model from five angles.
First, we check whether the information quotient separates normal, logical, and structural groups.
Second, we apply relation counterfactuals that preserve local token values but break their arrangement.
Third, we replace the fixed spatial hypergraph with random hypergraphs.
Fourth, we inspect which hyperedges contribute most to the anomaly score.
Fifth, we compare the information quotient with patch-kNN scores.
Together, these experiments test whether the model has learned category-specific normal structure rather than only a local matching function.

\subsection{Information-Quotient Separation}
Table~\ref{tab:iqgroups} and Figure~\ref{fig:iqgroups} summarize the information quotient on normal, logical, and structural groups.
Normal test images have a mean quotient of 0.6667.
Structural anomalies increase the mean quotient to 3.1861.
Logical anomalies increase it much more strongly, reaching 33.2278.
This pattern matches the design goal.
The quotient is not merely a generic defect score; it is most sensitive to the anomaly type that violates the learned normal-world structure.

The large gap between logical and normal images is also important conceptually.
Logical anomalies can contain normal local appearances.
If the score were only a nearest-normal-image mapping, the group separation would be weaker.
It would also be less aligned with relation violations.
Instead, the quotient rises most on the group whose abnormality is defined by world-rule violation.

\begin{table}[t]
\centering
\caption{Information quotient by group under the interpretability protocol.}
\label{tab:iqgroups}
\begin{tabular}{lrrrr}
\toprule
Group & N & Mean $\IQ$ & Std. & SEM \\
\midrule
Good test images & 102 & 0.6667 & 0.9439 & 0.0935 \\
Logical anomalies & 83 & \textbf{33.2278} & 130.9761 & 14.3765 \\
Structural anomalies & 90 & 3.1861 & 4.5737 & 0.4821 \\
\bottomrule
\end{tabular}
\end{table}

\begin{figure}[t]
\centering
\caption{Mean information quotient by group.
Logical anomalies produce the largest quotient, which indicates high explanation cost under the learned normal world.}
\label{fig:iqgroups}
\begin{tikzpicture}
\begin{axis}[
    ybar,
    width=0.74\linewidth,
    height=5.2cm,
    ymin=0,
    ymax=36,
    ylabel={Mean information quotient $\IQ$},
    symbolic x coords={Good,Structural,Logical},
    xtick=data,
    bar width=22pt,
    ymajorgrids=true,
    grid style={draw=gray!20},
    nodes near coords,
    every node near coord/.append style={font=\scriptsize},
    tick label style={font=\small},
    label style={font=\small},
]
\addplot[fill=methodteal!65, draw=methodteal] coordinates {
    (Good,0.6667)
    (Structural,3.1861)
    (Logical,33.2278)
};
\end{axis}
\end{tikzpicture}
\end{figure}

\begin{figure}[t]
\centering
\caption{t-SNE visualization of test images in the normal-world energy space.
Each point uses the six energy components of the proposed model.
Logical anomalies are separated from normal images, indicating that the learned energy signature contains category-specific world-structure information.}
\label{fig:tsne}
\begin{tikzpicture}
\begin{axis}[
    width=0.78\linewidth,
    height=6.0cm,
    xlabel={t-SNE dimension 1},
    ylabel={t-SNE dimension 2},
    xmin=-2.2, xmax=2.3,
    ymin=-2.3, ymax=2.4,
    grid=both,
    grid style={draw=gray!15},
    tick label style={font=\small},
    label style={font=\small},
    legend style={font=\small, draw=none, fill=none, at={(0.02,0.98)}, anchor=north west},
]
\addplot+[only marks, mark=*, mark size=1.4pt, color=methodgreen, opacity=0.65]
table[x=tsne_x,y=tsne_y,col sep=comma] {figures/tsne_good.csv};
\addlegendentry{Good}
\addplot+[only marks, mark=triangle*, mark size=1.7pt, color=methodred, opacity=0.75]
table[x=tsne_x,y=tsne_y,col sep=comma] {figures/tsne_logical.csv};
\addlegendentry{Logical}
\addplot+[only marks, mark=square*, mark size=1.5pt, color=methodblue, opacity=0.65]
table[x=tsne_x,y=tsne_y,col sep=comma] {figures/tsne_structural.csv};
\addlegendentry{Structural}
\end{axis}
\end{tikzpicture}
\end{figure}
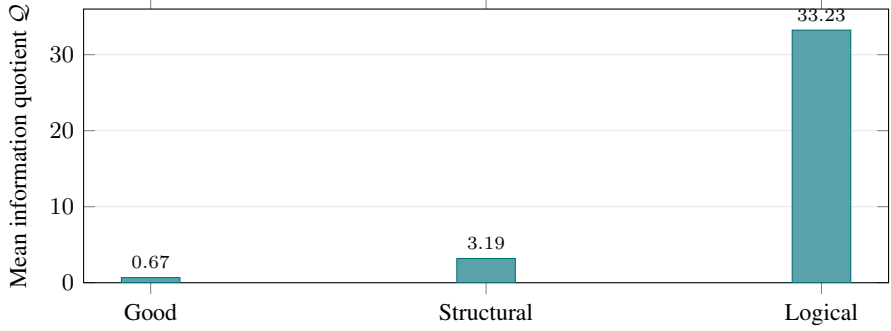

Figure~\ref{fig:tsne} provides a complementary view.
We embed the six-dimensional normal-world energy signature of each test image with t-SNE.
Each point is one test image, and colors indicate the ground-truth group.
The logical-anomaly points move away from the normal cluster in this learned energy space.
This supports the interpretation that the model forms a discriminative normal-world representation.
It is not only assigning a scalar score after a shallow mapping.

\begin{figure}[t]
    \centering
    \includegraphics[width=\linewidth]{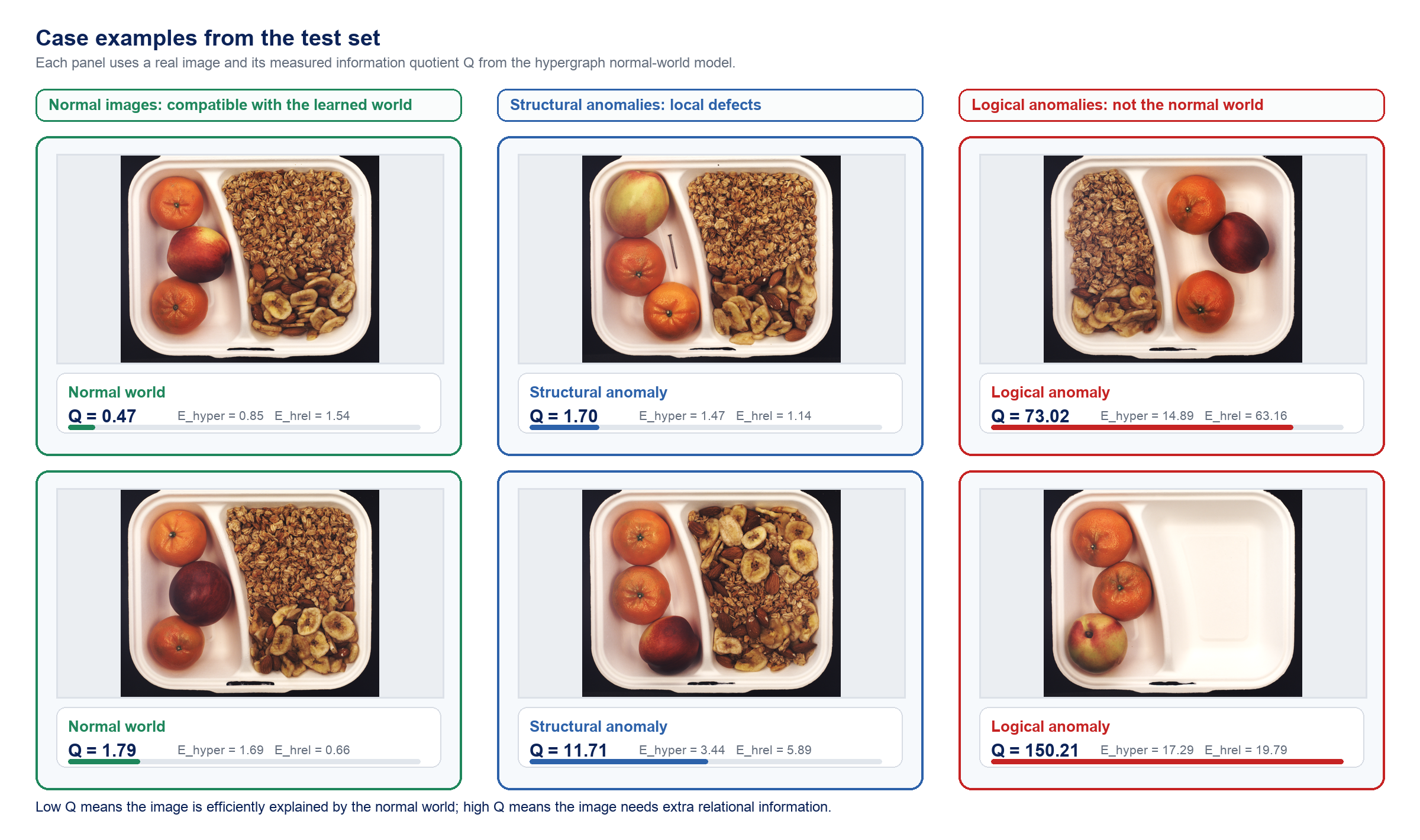}
    \caption{Qualitative case examples from the test set.
    Each panel shows a real image and its measured information quotient $\IQ$.
    Normal images have low scores, structural anomalies have moderate or high scores depending on local defects, and logical anomalies have very high scores because they are hard to explain under the learned normal world.}
    \label{fig:caseexamples}
\end{figure}

Figure~\ref{fig:caseexamples} shows qualitative examples from the same test set.
The normal images have low information quotients.
Structural anomalies can increase the quotient when local defects disturb patch and hyperedge evidence.
Logical anomalies produce much larger quotients, even though many visible parts remain plausible.
These examples make the score interpretation concrete.
Images that are not compatible with the learned normal world require much more relational information to explain.

\begin{figure}[p]
    \centering
    \includegraphics[width=0.92\linewidth]{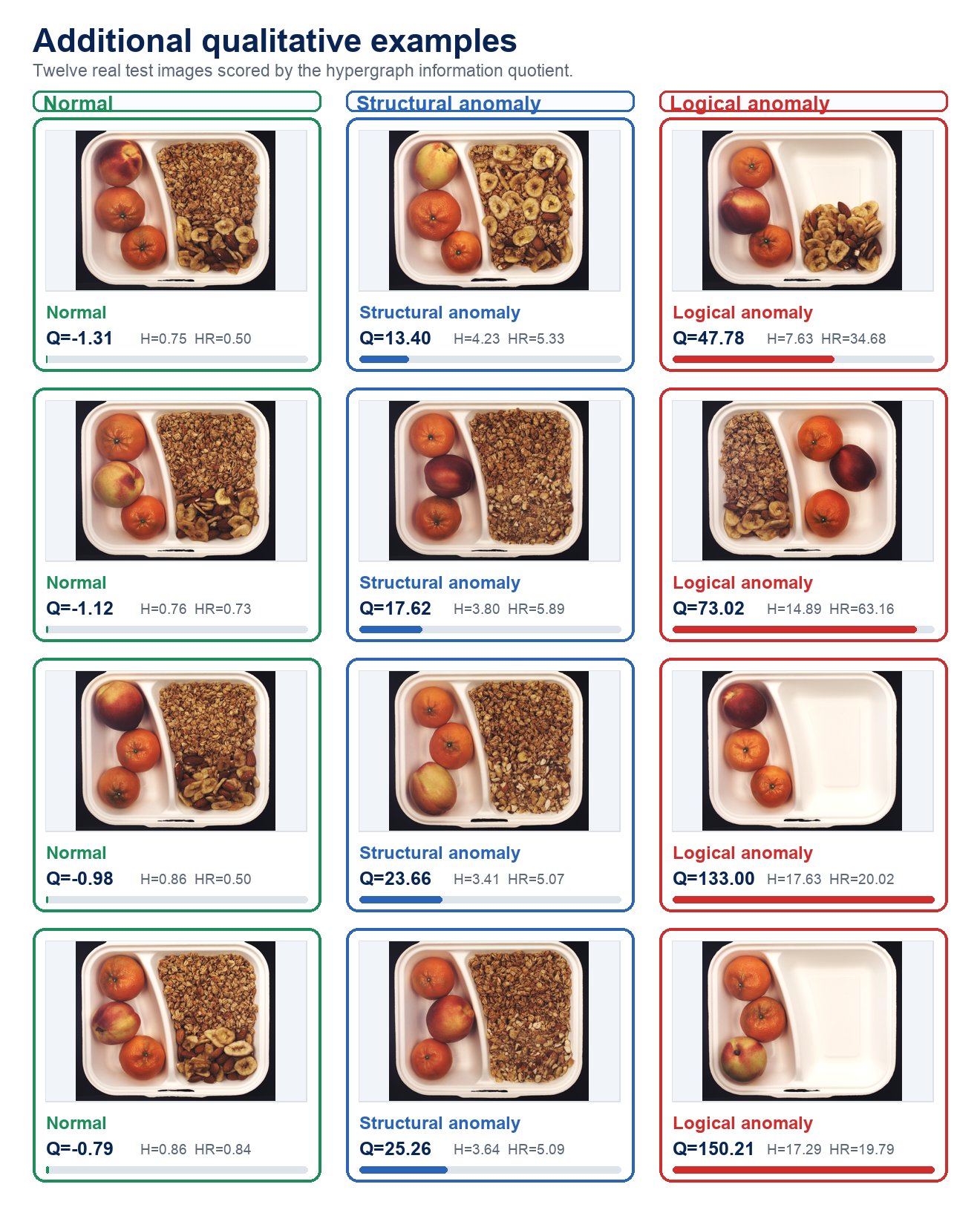}
    \caption{Additional qualitative examples.
    We show four normal images, four structural anomalies, and four logical anomalies.
    The pattern is consistent across samples: normal images are efficiently explained by the learned world, while logical anomalies often receive large hyperedge-relation scores because their object layout violates the normal relational state.}
    \label{fig:caseexamplesmore}
\end{figure}

\begin{figure}[p]
    \centering
    \includegraphics[width=0.92\linewidth]{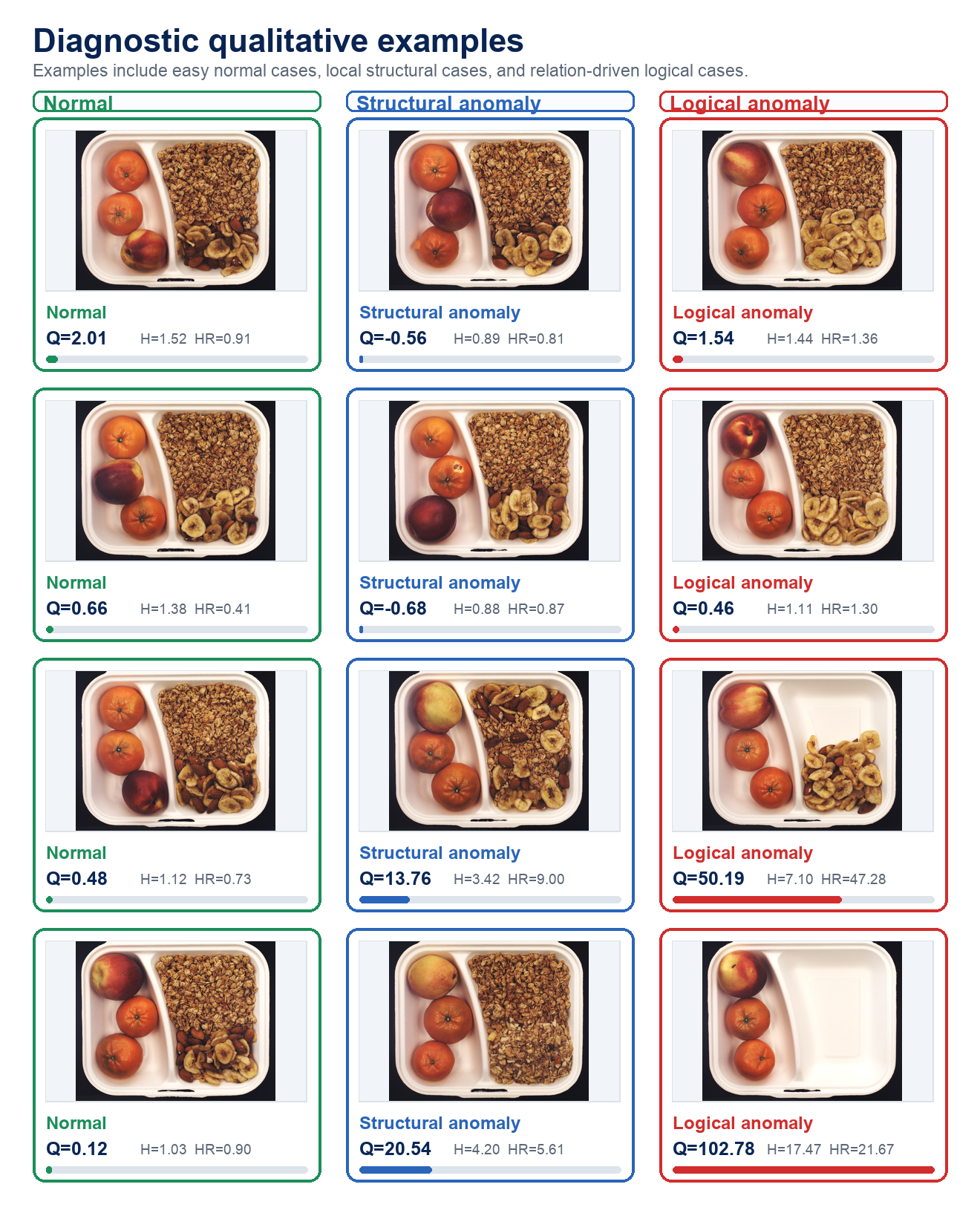}
    \caption{Diagnostic qualitative examples.
    Some structural cases remain relatively low because they are closer to local appearance defects, whereas logical cases become high when relation-bearing terms dominate.
    This supports the interpretation that the model is not only mapping images to anomaly scores, but also checking whether the image can be explained by the normal world structure.}
    \label{fig:caseexamplesdiagnostic}
\end{figure}

Figures~\ref{fig:caseexamples}--\ref{fig:caseexamplesdiagnostic} provide qualitative evidence for the same trend observed in the statistics.
The low-score normal panels preserve the learned compartment layout and object co-occurrence.
Structural anomalies can be detected when local evidence becomes unusual, but their scores are not always dominated by the relation terms.
Logical anomalies are different.
They often contain visually plausible food regions.
The hyperedge and hyperedge-relation terms become large because the global arrangement cannot be compressed into the learned normal world.

\subsection{Relation Counterfactual}
We take normal test images and permute token blocks.
This operation keeps many local token values in the image but breaks their spatial relations.
It is therefore a direct test of whether the model scores relation validity rather than only local appearance.
As shown in Table~\ref{tab:counterfactual}, the mean information quotient rises from 0.6667 to 83.8011.
The median increase is 85.0300, and every perturbed image increases its quotient.
The effect is not a small distributional fluctuation.
It is a consistent response to a controlled relation violation.

Figure~\ref{fig:counterfactual_energy} decomposes the counterfactual effect.
The largest increase comes from hyperedge-relation energy.
Patch-grid energy also increases, but it is much smaller.
The relation histogram does not change in this specific perturbation because the intervention preserves the global relation histogram used by that term.
This decomposition is exactly what we expect from a normal-world model.
When normal local evidence is rearranged into an invalid configuration, relation-bearing terms should dominate the score.

\begin{table}[t]
\centering
\caption{Relation counterfactual on normal test images.}
\label{tab:counterfactual}
\begin{tabular}{lrrrr}
\toprule
Experiment & Original $\IQ$ & Counterfactual $\IQ$ & $\Delta\IQ$ & Positive rate \\
\midrule
Block relation permutation & 0.6667 & 83.8011 & 83.1344 & 1.0000 \\
\bottomrule
\end{tabular}
\end{table}

\begin{figure}[t]
\centering
\caption{Energy decomposition under relation counterfactuals.
The hyperedge-relation term is the dominant source of the quotient increase, showing that the model reacts to broken normal relations.}
\label{fig:counterfactual_energy}
\begin{tikzpicture}
\begin{axis}[
    ybar,
    width=0.82\linewidth,
    height=5.6cm,
    ymin=0,
    ymax=62,
    ylabel={Mean energy increase},
    symbolic x coords={Patch grid,Relation hist.,Hyperedge,Hyperedge rel.},
    xtick=data,
    x tick label style={rotate=18, anchor=east, font=\small},
    bar width=16pt,
    ymajorgrids=true,
    grid style={draw=gray!20},
    nodes near coords,
    every node near coord/.append style={font=\scriptsize, rotate=90, anchor=west},
    label style={font=\small},
]
\addplot[fill=methodteal!65, draw=methodteal] coordinates {
    (Patch grid,7.0835)
    (Relation hist.,0.0000)
    (Hyperedge,18.8882)
    (Hyperedge rel.,57.4248)
};
\end{axis}
\end{tikzpicture}
\end{figure}

\subsection{Random Hypergraph Control}
If the hypergraph merely adds parameters, a random hypergraph should work as well.
Table~\ref{tab:random} shows the opposite.
The fixed spatial hypergraph reaches 0.9244 logical AUROC in this recomputed protocol.
Random hypergraphs range from 0.8982 to 0.9085.
Their mean logical AUROC is 0.9022.
This is below the fixed hypergraph by 0.0222 AUROC.

This control matters because it separates ``more features'' from ``meaningful structure''.
A random grouping still provides extra aggregate statistics, but those statistics are not aligned with stable spatial parts.
The drop shows that the benefit does not come from arbitrary pooling alone.
The fixed hypergraph encodes a useful inductive bias.
Normality is expressed through repeatable region groups and their relations.

\begin{table}[t]
\centering
\caption{Fixed hypergraph versus random hypergraph controls.}
\label{tab:random}
\begin{tabular}{lrrr}
\toprule
Variant & All AUROC & Logical AUROC & Structural AUROC \\
\midrule
Fixed hypergraph & 0.8157 & \textbf{0.9244} & \textbf{0.7154} \\
Random hypergraph, mean over 5 seeds & 0.7981 & 0.9022 & 0.7021 \\
\bottomrule
\end{tabular}
\end{table}

\subsection{Hyperedge Attribution}
We compute the largest hyperedge contribution for each group.
Logical anomalies have much larger top hyperedge contributions than normal images.
They also concentrate more of the anomaly score in the top three hyperedges.
This suggests that the score is localized in relation-bearing groups.
Table~\ref{tab:attrib} shows that the mean top-1 contribution is 3.9148 for normal images, 8.9302 for structural anomalies, and 46.2437 for logical anomalies.
The top-three share also rises from 0.1613 on normal images to 0.2416 on logical anomalies.

This attribution result gives a different type of evidence from AUROC.
The score is not only numerically higher for logical anomalies.
It is higher because a small number of region groups become hard to explain.
This is what a world-model violation should look like.
A specific part of the visual world no longer fits the normal relational state.

\begin{table}[t]
\centering
\caption{Hyperedge attribution summary.}
\label{tab:attrib}
\begin{tabular}{lrrrr}
\toprule
Group & N & Top-1 contribution & Top-3 share & Mean $\IQ$ \\
\midrule
Good & 102 & 3.9148 & 0.1613 & 0.6667 \\
Logical anomalies & 83 & \textbf{46.2437} & \textbf{0.2416} & \textbf{33.2278} \\
Structural anomalies & 90 & 8.9302 & 0.1888 & 3.1861 \\
\bottomrule
\end{tabular}
\end{table}

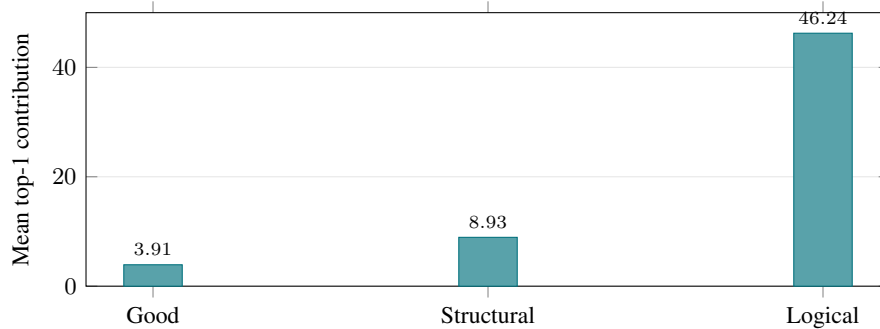
\begin{figure}[t]
\centering
\caption{Hyperedge attribution strength by group.
Logical anomalies concentrate much larger contributions in the most responsible hyperedge.}
\label{fig:attrib}
\begin{tikzpicture}
\begin{axis}[
    ybar,
    width=0.74\linewidth,
    height=5.2cm,
    ymin=0,
    ymax=50,
    ylabel={Mean top-1 contribution},
    symbolic x coords={Good,Structural,Logical},
    xtick=data,
    bar width=22pt,
    ymajorgrids=true,
    grid style={draw=gray!20},
    nodes near coords,
    every node near coord/.append style={font=\scriptsize},
    tick label style={font=\small},
    label style={font=\small},
]
\addplot[fill=methodteal!65, draw=methodteal] coordinates {
    (Good,3.9148)
    (Structural,8.9302)
    (Logical,46.2437)
};
\end{axis}
\end{tikzpicture}
\end{figure}

\subsection{Texture Versus Relation Interventions}
We compare a small local token noise intervention with the relation permutation.
Texture noise changes the information quotient by only 0.1612 on average.
Relation permutation changes it by 83.1344.
The largest component change comes from hyperedge-relation energy.
This experiment is a controlled comparison between local appearance perturbation and relation perturbation.
Both interventions are applied to normal images.
The texture intervention slightly changes local evidence while preserving the world layout.
The relation intervention preserves many token values but changes their configuration.
The model reacts far more strongly to the second intervention.

This result supports the claim that the learned score is relation-aware.
If the method were only measuring local feature noise, the texture intervention should have been competitive.
Instead, the relation intervention is more than two orders of magnitude stronger in $\Delta\IQ$.

\begin{table}[t]
\centering
\caption{Texture-only versus relation-only interventions.}
\label{tab:interventions}
\begin{tabular}{lrrrr}
\toprule
Intervention & Original $\IQ$ & New $\IQ$ & $\Delta\IQ$ & Positive rate \\
\midrule
Local texture noise & 0.6667 & 0.8279 & 0.1612 & 0.9902 \\
Relation permutation & 0.6667 & 83.8011 & 83.1344 & 1.0000 \\
\bottomrule
\end{tabular}
\end{table}

\subsection{Comparison with Patch-kNN Scores}
We compare the information quotient with the patch-kNN distance.
The two scores are only moderately correlated overall.
On structural anomalies, Pearson correlation is 0.1127.
This confirms that patch-kNN and the information quotient capture different mechanisms.
The overall Pearson correlation is 0.4968 and the overall Spearman correlation is 0.6098.
This means the two scores are related, as both use visual features, but they are not the same detector.
The correlation is high on logical anomalies, where both methods often assign high scores to difficult cases.
It is much lower on normal and structural samples.
In particular, structural Pearson correlation is only 0.1127.
This supports the complementarity observed in Table~\ref{tab:main}.
Patch-kNN is better for local structural defects.
The information quotient adds a different relation-sensitive signal.

Taken together, the five analyses support the normal-world interpretation.
The score separates logical anomalies, rises under relation counterfactuals, depends on meaningful hypergraph structure, localizes to hyperedges, and is not reducible to patch-kNN distance.
These findings are difficult to explain by a simple image-to-score mapping alone.
They are consistent with the view that the model has distilled category-specific world knowledge from normal images.

\begin{table}[t]
\centering
\caption{Correlation between information quotient and patch-kNN distance.}
\label{tab:corr}
\begin{tabular}{lrrr}
\toprule
Scope & N & Pearson & Spearman \\
\midrule
All samples & 688 & 0.4968 & 0.6098 \\
Good test images & 102 & 0.2972 & 0.2773 \\
Logical anomalies & 83 & 0.8673 & 0.8520 \\
Structural anomalies & 90 & 0.1127 & 0.4688 \\
\bottomrule
\end{tabular}
\end{table}

\section{Discussion and Limitations}
The experiments support a specific claim.
The proposed model is not a universal replacement for patch memory.
It is a relation-aware normal-world score.
Patch-kNN is stronger on structural anomalies in our current experiments.
Our model is stronger on logical anomalies, and the interpretability experiments show that it responds sharply to relation violations.

The current study has limitations.
First, only the breakfast-box category is available in the current extracted dataset.
A complete benchmark study should run all five MVTec LOCO categories and additional industrial datasets such as MVTec AD.
Second, the hypergraph is fixed and hand designed.
Future work should learn category-adaptive hyperedges or use object-part segmentation.
Third, a complete industrial model should combine local patch memory and normal-world relation modeling, because the two signals capture different failure modes.

\section{Conclusion}
We proposed the \method\ for normal-only logical visual anomaly detection.
The method distills a normal visual world from frozen DINOv2 patch tokens and measures test images with a hypergraph information quotient.
On the available MVTec LOCO breakfast-box validation data, the model achieves the best logical anomaly AUROC among reproduced baselines and ablations.
Five interpretability experiments further show that the score is sensitive to relation violations, fixed spatial hyperedges, and hyperedge-level contributions.
The central takeaway is simple.
Visual anomaly detection should model normal relations, not only normal patches.

\bibliographystyle{plain}
\bibliography{references}

\end{document}